\documentclass{article}

\usepackage[preprint]{neurips_2023}

\usepackage[utf8]{inputenc} 
\usepackage[T1]{fontenc}    
\usepackage{hyperref}       
\usepackage{url}            
\usepackage{multirow}
\usepackage{booktabs}       
\usepackage{amsfonts}       
\usepackage{amsmath}
\usepackage{nicefrac}       
\usepackage{microtype}      
\usepackage{xcolor}         
\usepackage{graphicx}

\newtheorem{model}{Model}

\title{Voice Biomarkers for Depression and Anxiety}

\author{%
  Oleksii Abramenko\thanks{Equal contribution. This work was performed while the authors were the machine learning team at Kintsugi Mindful Wellness, Inc.
  Please refer to \textit{\nameref{sec:acknowledgements}} section for more details.} \\
  \texttt{oleksii.abramenko.90@gmail.com} \\
  \And
  Noah D.~Stein\footnotemark[\value{footnote}] \\
  \texttt{nstein@mit.edu} \\  
  \And
  Colin Vaz\footnotemark[\value{footnote}]\\
  \texttt{colinvaz@gmail.com}
}

\begin{document}

\maketitle

\begin{abstract}
Current approaches to detecting depression and anxiety from speech primarily rely on machine learning techniques that utilize
hand-engineered paralinguistic features and related acoustic descriptors derived
from time- and frequency-domain representations of speech signals. Applying deep learning methods directly to
raw speech signals has the potential to produce biomarker representations
with substantially greater predictive power. However, these approaches typically require large volumes of carefully annotated data
to learn robust and clinically meaningful representations of the underlying biomarkers.
In this paper, we describe our efforts toward developing a deep learning model trained on a large-scale
proprietary dataset comprising  \textasciitilde 65,000 utterances collected from more than 23,000 subjects
representative of relevant United States demographics. We present the techniques employed and analyze their impact on model performance.
Our results demonstrate that the proposed models can extract content-agnostic biomarker information, which, when combined with lexical features extracted from audio,
yields improved predictive performance in production settings. Our models are evaluated on \textasciitilde 5000 unique subjects and achieve performance of 71\%
in terms of sensitivity and specificity. To foster further research in mental health assessment from speech,
we release the best-performing model described in this paper on HuggingFace.
\end{abstract}

\section{Introduction}
In the United States, depression affects millions of adults each year and remains one of the leading contributors to disability (\citet{hdrf_depression_facts}) and reduced quality of life (\citet{nimh_major_depression}).
Despite growing public awareness surrounding mental health, the identification and treatment of depression continue to face significant barriers,
including social stigma, underreporting of symptoms, and limited access to consistent screening practices.
These challenges are particularly pronounced among older adults, who are less likely to seek mental health support and often experience depression alongside
other chronic health conditions (\citet{nndc_facts}). As the aging population in the United States continues to expand (\citet{apa_stigma}), the burden of untreated mental health conditions
among older individuals is expected to increase substantially. At the same time, current screening approaches rely heavily on self-reported questionnaires
and subjective clinical evaluations, creating a need for more objective, accessible, and scalable methods for detecting depressive symptoms.
Developing reliable tools for early identification of depression may help improve intervention strategies, reduce long-term health complications,
and support better outcomes for vulnerable populations.

In this paper we present our work towards building speech-based model designed to screen for signs of depression and anxiety using voice biomarkers.
To the best of our knowledge, it is the first model developed explicitly for clinical-grade mental health assessment from speech without
reliance on linguistic content or transcription. A predecessor model has been peer-reviewed in the largest voice biomarker study by
the Annals of Family Medicine (\citet{Mazur240091}). The model operates exclusively on the acoustic properties of the speech signal,
extracting depression- and anxiety-specific voice biomarkers rather than semantic or lexical information.
Numerous studies, such as \citet{ALMAGHRABI2023105020}, \citet{mundt2012vocal} and \citet{menne2024voice} have demonstrated that paralinguistic features – such as spectral entropy, pitch variability, fundamental frequency, and
related acoustic measures – exhibit strong correlations with depression and anxiety. Building on this body of evidence, our model extends prior approaches
by leveraging deep learning to learn fine-grained vocal biomarkers directly from the raw speech signal, yielding representations that
demonstrate greater predictive power than hand-engineered paralinguistic features. DAM analyzes spoken audio to estimate depression and anxiety severity
scores which can be subsequently mapped to standardized clinical scales, such as PHQ-9 (Patient Health Questionnaire-9) for depression and
GAD-7 (Generalized Anxiety Disorder-7) for anxiety.

The paper is structured as follows. Section~\ref{sec:settings} outlines the problem statement, describes the available data and its demographics, specifies the model architecture and evaluation metrics.
Section~\ref{sec:experiments} presents the experiments conducted in conjunction with the methods used to improve model performance, detailing the techniques applied and quantifying their impact.
Finally, Section~\ref{sec:discussion} discusses our findings, including model characteristics and some prominent biomarker properties.

\section{Setting}
\label{sec:settings}
\subsection{Labels}
\label{sec:labels}
In this paper we consider models for predicting two types of mental health labels:
\begin{itemize}
\item The PHQ-$9$ \citet{phq9} is a widely-used self-administered depression survey.
It asks the subject to answer $9$ questions quantifying frequency of depression-related symptoms (changes in energy, sleep, appetite, etc.) over the past two weeks.
Each question is answered on a scale from $0$ (not at all) to $3$ (nearly every day).
Unless otherwise specified, results are summarized by summing the questions' scores to get a total score $0-27$.
Key total score ranges are $0-9$ for minimal to mild depression, $10-14$ for moderate depression, and $15-27$ for severe depression.
\item The GAD-$7$ \citet{gad7} is a self-administered anxiety survey, with $7$ questions answered on the same $0-3$ scale and summed.
Key total score ranges are $0-4$ for minimal anxiety, $5-9$ for mild anxiety, $10-14$ for moderate anxiety, and $15-21$ for severe anxiety.
\end{itemize}


\subsection{Problem statement}
Our primary goal is building models to screen people for anxiety and depression based on speech audio.
To enable broad applicability, no constraints are placed on the content of the speech; in particular it is not expected to directly relate to mental health.

More specifically, the models take $30$ seconds or more of speech and produce a floating point score for each of depression and/or anxiety.
The scores are intended to be monotonically correlated with the PHQ-$9$ sum 
for depression and the GAD-$7$ sum for anxiety, in the sense that higher scores tend to be associated with higher labels (in a possibly nonlinear way).

As a secondary goal we consider models taking text transcribed from speech as input, either instead of or in addition to audio, with the same outputs and labels.
In the motivating application transcription was avoided for privacy reasons.

\subsection{Data}
\label{sec:data}

Each data sample used for training and evaluating models in this paper consists of an audio stream, a PHQ-9 label, and a GAD-7 label.

Each audio stream is a 16 kHz mono recording of a single talker, the person to whom the above labels apply.
Participants were given a choice of several prompts unrelated to mental health to talk about, such as a favorite holiday, type of music, sport, or actor.
They were instructed to talk on the subject in English for $1-3$ minutes, avoiding background noise.
Data collection was done in a distributed fashion, with recordings made on the participants' own phones, tablets, and laptops.
Some participants recorded answers to multiple prompts.

The participants were recruited from across the United States, representing all $50$ states.
The data have been divided into speaker-disjoint train, validation, and test splits which are roughly demographically balanced (Table~\ref{tab:datastats}).

\begin{table}[!ht]
\caption{Statistics for PHQ-9 and GAD-7 labeled data.}
\label{tab:datastats}
\tiny
    \centering
    \begin{tabular}{llrrrr}
    \toprule
        ~ & ~ & Train & Validation & Test & Total \\
        \midrule
        \multirow{2}{*}{Counts} & Unique subjects & 23,743 & 6,047 & 5,353 & 34,457 \\ 
        & Recordings & 43,945 & 11,073 & 9,810 & 64,828 \\
        \midrule
        \multirow{4}{*}{Audio duration} & Mean (std.\ dev.) (s) & 56.4 (13.6) & 56.8 (13.8) & 56.1 (13.6) & 56.4 (13.7) \\
        & Median (s)& 56.8 & 57.6 & 55.7 & 56.7 \\
        & Range (s) & 30-229 & 31-231 & 31-176 & 30-231 \\
        & Total (hr)& 688.5 & 174.7 & 152.9 & 863.2 \\
        \midrule
        \multirow{3}{*}{Age (yr)}& Range & 18+ & 18+ & 18+ & 18+ \\
        & Mean (std.\ dev.) & 45.2 (16.8) & 45.4 (16.8) & 45.5 (16.7) & 45.3 (16.8) \\
        & Median & 42 & 43 & 43 & 42 \\
        \midrule
        \multirow{4}{*}{Gender, \%} & Female & 57.5 & 57.8 & 57.8 & 57.6 \\
       &  Male & 41.1 & 41 & 40.7 & 41 \\
        & Not specified & 0.4 & 0.4 & 0.3 & 0.3 \\
       &  Other & 0.8 & 0.7 & 0.8 & 0.8 \\
        \midrule
        \multirow{7}{*}{Race/ethnicity, \%} &  Asian or Pacific Islander & 5.9 & 6.0 & 6.0 & 6.0 \\
        & Black or African American & 12.6 & 12.5 & 11.8 & 12.5 \\
        & Hispanic or Latine & 17.0 & 17.2 & 17.7 & 17.2 \\
        & Native American or American Indian & 0.9 & 0.8 & 0.8 & 0.8 \\
        & Not specified & 2.9 & 2.9 & 2.9 & 2.9 \\
        & Other or mixed race & 4.4 & 4.4 & 4.4 & 4.4 \\
        & White & 56.2 & 56.1 & 56.5 & 56.2 \\
        \midrule
        \multirow{4}{*}{PHQ-9 score} & Mean (std.\ dev.) & 8.7 (6.7) & 8.7 (6.8) & 8.8 (6.8) & 8.7 (6.7) \\
        & Median & 7 & 7 & 8 & 7 \\
        & Mode & 0 & 0 & 0 & 0 \\
        & Range & 0-27 & 0-27 & 0-27 & 0-27 \\
        \midrule
        \multirow{4}{*}{GAD-7 score} & Mean (std.\ dev.) & 7.3 (6.1) & 7.3 (6.2) & 7.4 (6.2) & 7.3 (6.2) \\
        & Median & 6 & 6 & 6 & 6 \\
        & Mode & 0 & 0 & 0 & 0 \\
        & Range & 0-21 & 0-21 & 0-21 & 0-21 \\
        \bottomrule
    \end{tabular}
\end{table}



\subsection{Architectures}
\label{sec:architectures}

All depression and anxiety screening models in this paper consist of the following pieces:
\begin{enumerate}
\item One or more parallel pre-trained backbones, each with
\begin{enumerate}
\item A fixed feature extractor or tokenizer
\item The frozen backbone model itself
\item Trainable LoRA \citet{lora} modules to adapt the backbone's linear and attention layers
\begin{enumerate}
\item Rank $32$
\item $\alpha = 64.0$
\item Dropout (40\%)
\end{enumerate}
\item Backbone-specific temporal pooling
\end{enumerate}
\item A fully connected network taking all pooled backbone features as input
\begin{enumerate}
\item One hidden layer of size $256$
\item Mish nonlinearity \citet{mish}
\item Output layer of size $64$
\end{enumerate}
\item A separate fully connected head for each task (depression and anxiety) taking these common features as input
\begin{enumerate}
\item One hidden layer of size $128$
\item Mish nonlinearity
\item Dropout (40\%)
\item A scalar output layer
\end{enumerate}
\end{enumerate}
Where not otherwise specified, all audio backbones are \texttt{openai/whisper-small.en} \cite{whisper}.
The associated temporal pooling of backbone features is mean pooling.
Since this model has a $30$ second receptive field, inputs during training are randomly selected $30$-second segments.
Inputs during validation and testing are all $30$-second segments, the final one padded with silence, with the output scores averaged over segments.

By default, text backbones are \texttt{google-bert/bert-base-uncased} \cite{bert}, with the ``pooled'' output being the output corresponding to the \texttt{[CLS]} token. The input is text transcribed from the spoken prompt recordings using \texttt{openai/whisper-large}.

\subsection{Metrics}
\label{sec:metrics}
We want models which are able to detect each condition (depression and anxiety) at various severity levels.
For each condition, we compute metrics for several binary decision problems based on the key score thresholds in \ref{sec:labels}.
For depression we consider the problems of predicting PHQ-9 $<k$ vs.\ PHQ-9 $\geq k$ for $k=10, 15$ and for anxiety the problems of predicting GAD-7 $<k$ vs.\ GAD-7$\geq k$ for $k=5, 10, 15$.

For each condition and associated decision problem, we compute optimal thresholds and associated metrics.
Our primary metric is the highest simultaneously achievable sensitivity and specificity, expressed as a percentage and denoted $S_n=S_p$.
This is equivalently $100\%$ minus the Equal Error Rate.
For each condition the metrics are averaged over the associated decision problems to produce an aggregate metric for that condition, also denoted $S_n=S_p$.
Models with higher $S_n=S_p$ are deemed better in aggregate.

Similarly, for each decision problem an Area Under the ROC curve is computed at these areas are also averaged over the decision problems.

\section{Modeling  Techniques}
\label{sec:experiments}

In this section, we present some of the approaches we took to improve the model's performance and depression and anxiety $S_n=S_p$. We present these techniques in the chronological order that we tried them, so the models build upon each other. The results in each subsection provide a rough snapshot of model performance.
For each experiment, the epoch at which the validation $S_n=S_p$ on the depression task is maximized is chosen and the test $S_n=S_p$ is computed.
The depression task is chosen for business reasons, though the depression and anxiety performances tend to correlate well.

\subsection{Audio backbones}
\label{sec:audio_backbones}
We experimented with using different pre-trained audio models as a backbone for depression and anxiety classification. We explored models that were used for automatic speech recognition (ASR) (Whisper Small and Base encoders \citet{whisper}
) and speaker verification (Titanet Large \citet{titanet} and ECAPA-TDNN \citet{ecapa_tdnn}). We chose these models because they are trained on a wide variety of speakers, so leverage the high-level speech representation captured by these models. Furthermore, the model sizes made experimentation tractable on Nvidia T4 GPUs. A classification head, as described in Section~\ref{sec:architectures}, is attached to each backbone and fine-tuned for depression classification. Table~\ref{tab:audio_backbones} shows the AUC after fine-tuning each backbone, and also includes information about the backbone sizes and approximate amount of data used to pre-train the backbones.

\begin{table}[h]
	\centering
	\caption{Specifications (\# of parameters and amount of data used for training) of the audio backbones we tested, and AUC after fine-tuning the backbones for depression classification.}
	\label{tab:audio_backbones}
	\begin{tabular}{|l|l|l|l|}
		\hline
		\textbf{Backbone} & \textbf{\# params} & \textbf{Training data amount (hours)} & \textbf{Depression AUC} \\
		\hline
		Whisper Small & 80 M & 680 K & 0.687\footnotemark[\value{footnote}] \\
		Whisper Base & 20 M & 680 K & 0.668\footnotemark[\value{footnote}] \\  
		Titanet Large & 25 M & 4 K & 0.640\footnotemark[\value{footnote}] \\  
		ECAPA-TDNN & 20 M & 4 K & 0.630\footnotemark[\value{footnote}] \\  
		\hline
	\end{tabular}
\end{table}
\footnotetext{Based on an earlier model architecture and dataset; directly comparable within Section~\ref{sec:audio_backbones} but not to numbers in other sections.}
As can be seen in Table~\ref{tab:audio_backbones}, there is a strong correlation between the pre-trained model size and the depression AUC, with larger models leading to higher AUC. However, the amount of data used for pre-training also has a notable effect on the AUC; one can see that Whisper Base has around the same number of parameters as Titanet Large and ECAPA-TDNN but over 100x data used for pre-training, and this leads to almost $0.03$ higher AUC. This finding suggests that using a model that is pre-trained on a larger amount of data from a wider range of speakers is beneficial when fine-tuning for depression classification. From the results of the audio backbone investigation, we found Whisper Small to be the best audio backbone for depression classification.

\subsection{Ordinal regression (OR)}
\label{sec:or}
PHQ-9 and GAD-7 total scores are often used as measures of severity of depression and anxiety.
It is worthwhile to have a single model which can simultaneously solve multiple related decision problems, such as binary classification of severe depression; binary classification of moderate or higher anxiety; quaternary classification of anxiety into severity buckets; or ternary classification of depression into positive, negative, or ``indeterminate''.
A natural framework for this is ordinal regression, which encourages a model to make its scalar output monotonically related to some linearly ordered labels, such as PHQ-9 or GAD-7 scores.

We adopt the \textbf{CORAL loss} \citet{coral}.
For a dataset $\mathcal{D}$ of pairs $(x, y)$ with $y\in\{0, 1, \ldots, n\}$ and scalar-valued network $s_\theta$ parametrized by $\theta$, minimizing the CORAL loss tries to make $s_\theta(x)$ a score which works well to simultaneously solve all the binary decision problems $y < k$ vs.\ $y \geq k$, with decision threshold depending on $k$.
To do so the CORAL loss sums binary cross entropy over all such decision problems $k=1, \ldots, n$, with an independent trainable bias $b_k$ for each decision problem:
\begin{equation*}
\operatorname{CORAL}(x, y; \theta, b_1, \ldots, b_n) = -\sum_{k=1}^y\log\left(s_\theta(x) + b_k\right) - \sum_{k=y+1}^n\log\left(1 - s_\theta(x) - b_k\right).
\end{equation*}
The biases $b_1, \ldots, b_n$ are trained jointly with the neural net parameters $\theta$.
A scalar bias at the final layer of $s_\theta$ would be redundant and can be omitted.
As usual, this loss is optimized by mini-batch SGD or a variation thereof, in this case AdamW \citet{adamw}.

Given a collection of pairs of scores and labels $(s_\theta(x), y)$, there exist dynamic programming methods to find optimal thresholds under many natural decision criteria for a variety of related binary and multi-class classification problems, such as those mentioned above.
See for example \citet{parallel_or_dp} or the implementations used for the present application at \url{https://huggingface.co/KintsugiHealth/dam/tree/main/tuning}.


\begin{model}
\label{model:or}
Training a model with a single LoRA-adapted \texttt{whisper-small.en} backbone and heads for depression and anxiety, we achieve $67.5\%$ and $67.7\%$ test $S_n=S_p$, respectively for the two tasks, and $0.740$ and $0.746$ AUROC.
The model achieves its maximum performance level after a few epochs and then quickly overfits (Figure~\ref{fig:svleffect}).
This behavior is typical, with batch size, learning rate, LoRA, and dropout hyperparameter tuning affecting the time to best performance, but not the performance level reached (details omitted).
\end{model}

\subsection{Text models}
\label{sec:text}
As with the audio backbones, we also experimented with using different pre-trained text models as a backbone for depression and anxiety classification. The idea was to extract lexical information that can be complementary to the acoustic information extracted from the audio backbone. We explored encoder-only models (BERT \citet{bert} and RoBERTa \citet{roberta}), a decoder-only model (GPT-2 \citet{gpt2}), and an encoder-decoder model (T5 \citet{t5}). Table~\ref{tab:text_backbone_specs} shows some specifications of these models.

\begin{table}[h]
	\centering
	\caption{Number of parameters and amount of data used for training various pre-trained text models.}
	\label{tab:text_backbone_specs}
	\begin{tabular}{|l|l|l|}
		\hline
		\textbf{Backbone} & \textbf{\# params} & \textbf{Training data size (GB)} \\
		\hline
		BERT Base & 110 M & 16 \\
		BERT Large & 340 M & 16 \\
		RoBERTa Base & 110 M & 161 \\
		GPT-2 & 124 M & 40 \\
		T5 Base & 220 M & 750 \\
		\hline
	\end{tabular}
\end{table}


The audio for all subjects was transcribed using Whisper Large, and the transcriptions were input into each pre-trained model. For BERT and RoBERTa models, we extract embeddings from the last layer for the \texttt{[CLS]} token that is prepended to start of the input sequence. For GPT-2, we obtain embeddings from the last layer for the last token in the sequence. For T5, we first prepend "mental: " to each transcription to serve as the task token, and we extract embeddings from the last layer for the task token. The text embeddings are optionally concatenated with the audio embeddings from Whisper Small and fed into a depression and anxiety classification network. Table~\ref{tab:text_backbones} shows the $S_n=S_p$ and AUC after fine-tuning each text backbone (the Whisper Small backbone is also fine-tuned simultaneously).

\begin{table}[h]
	\centering
	\caption{$S_n=S_p$ and AUC after fine-tuning various text backbones along with Whisper Small audio backbone for depression and anxiety classification.}
	\label{tab:text_backbones}
	\begin{tabular}{|l|l|l|l|l|}
		\hline
		\textbf{Task} & \textbf{Text backbone} & \textbf{Audio backbone} & $S_n=S_p$ & \textbf{AUC} \\
		\hline
		\multirow{5}{*}{Depression} & BERT Base & None & $68.3$ & $0.747$ \\ 
		& BERT Base & Whisper Small & $70.0$ & $0.750$ \\  
		& BERT Large & Whisper Small & $70.1$ & $0.774$ \\  
		& RoBERTa Base & Whisper Small & $69.6$ & $0.780$ \\  
		& GPT-2 & Whisper Small & $69.3$ & $0.765$ \\  
		& T5 & Whisper Small & $68.8$ & $0.759$ \\  
		\hline
		\multirow{5}{*}{Anxiety} & BERT Base & None & $67.2$ & $0.736$ \\
		& BERT Base & Whisper Small & $69.9$ & $0.769$ \\
		& BERT Large & Whisper Small & $69.9$ & $0.769$ \\
		& RoBERTa Base & Whisper Small & $69.9$ & $0.776$ \\
		& GPT-2 & Whisper Small & $69.4$ & $0.764$ \\
		& T5 & Whisper Small & $68.7$ & $0.759$ \\
		\hline
	\end{tabular}
\end{table}

From Table~\ref{tab:text_backbones}, one can see that the metrics are very similar across the backbones, with encoder-only backbones (BERT and RoBERTa) performing slightly better than GPT-2 and T5. We hypothesize that the higher performance with the BERT-based backbones is due to the way these models are trained, where the class token learns to capture semantics and summarize high-level information of the entire text. Meanwhile, decoder-only and encoder-decoder models, like GPT-2 and T5 are trained for next token prediction rather than summarization. Of note, there is negligible difference in metrics between BERT Base, BERT Large, and RoBERTa Base; this finding suggests that our downstream task did not benefit from a larger backbone (as in the case of BERT Large) or a similar sized backbone that is trained on more data (as in the case of RoBERTa). So we decided to use BERT Base as the text backbone.

\begin{model}[Text-only]
\label{model:text}
The best model taking only transcribed text as input with no audio backbone achieved $68.3\%$ $S_n=S_p$ and $0.747$ AUC for depression (resp.\ $67.2\%$ and $0.736$ for anxiety). This model is used as part of the teacher Model~\ref{model:kdteacher} in the knowledge distillation process for training student Model~\ref{model:kdstudent}.
\end{model}

\subsection{Score variance loss (SVL)}
\label{sec:svl}

\begin{figure}[tbp]
\begin{center}
\includegraphics[width=0.8\textwidth]{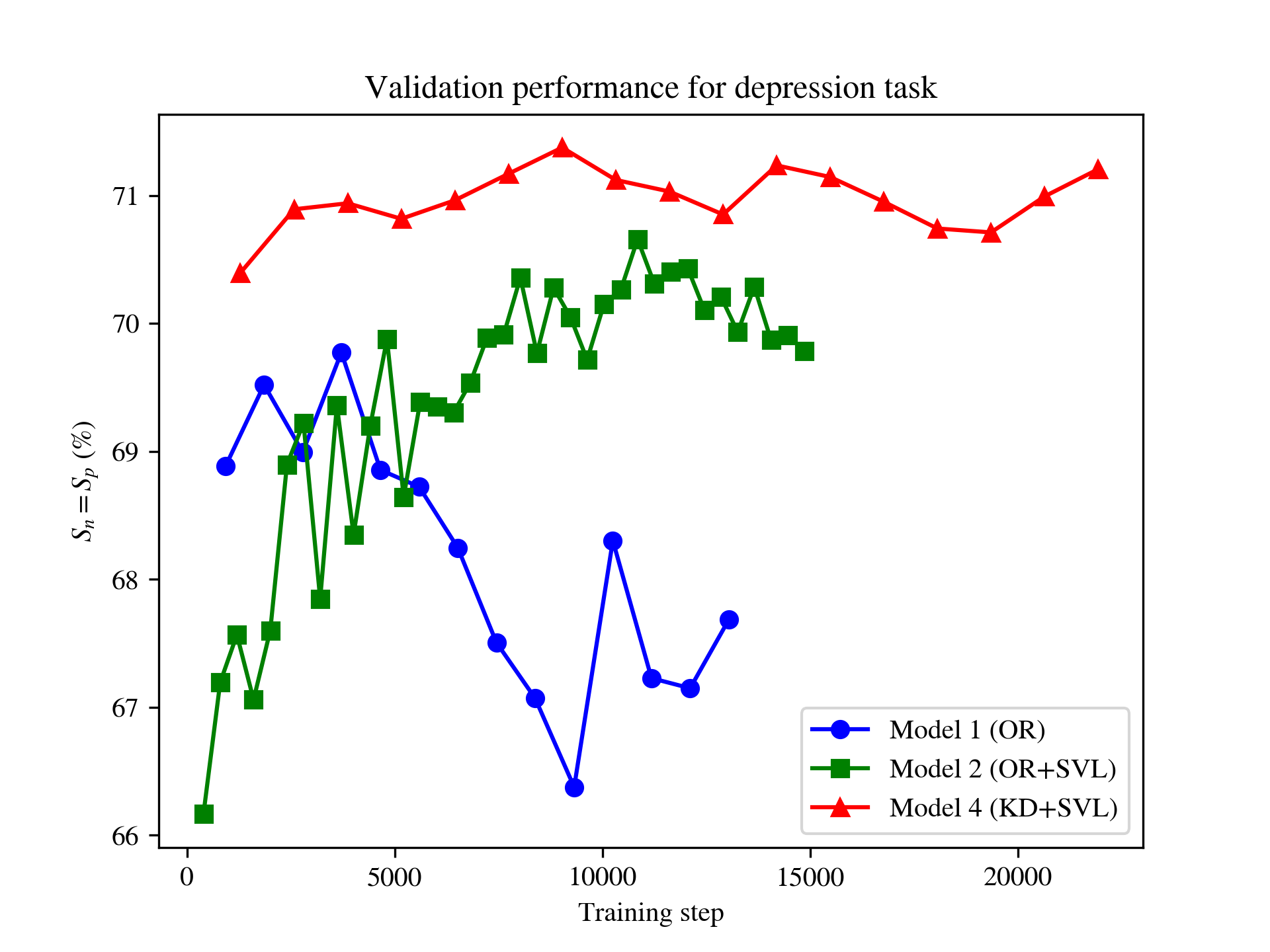}
\caption{Validation $S_n=S_p$ for audio-only models on the depression task with ordinal regression, score variance loss, and knowledge distillation.}
\label{fig:svleffect}
\end{center}
\end{figure}

The best method we have found for avoiding this early overfitting and improving the final performance is to encourage score consistency across examples from the same voice.
This regularizes the effect of label noise \citet{crcirln, udact}, which is expected based on the personal and subjective nature of the PHQ-$9$ and GAD-$7$ questions.
In our datasets, all audio clips from the same voice are recorded within a few minutes of each other and share PHQ-$9$ and GAD-$7$ labels, so score consistency between samples is a natural condition to impose.

To do so, we form training mini-batches by selecting a number of voices (here $128$) and two or more (here two) audio clips per voice.
The model produces a scalar score for each audio clip as before.
In addition to computing the CORAL loss from each score, we compute the variance of the scores for each voice and average over voices.
In the case of two audio training examples $(x_1, y), (x_2, y)$ for the same voice, this \textbf{score variance loss} is just
\begin{equation*}
\operatorname{SVL}(x_1, x_2, y; \theta) = \frac{1}{4}\bigl(s_\theta(x_1) - s_\theta(x_2)\bigr)^2,
\end{equation*}
independent of the shared label $y$.
Minimizing this quantity encourages consistency $s_\theta(x_1) \approx s_\theta(x_2)$.
The SVL is in tension with the CORAL loss; the former can be optimized in isolation by making $s_\theta$ a constant function, which is presumably not optimal for the latter.

\begin{model}
\label{model:svl}
We optimize a linear combination of CORAL and SVL via SGD, using weighting the losses in the ratio $1:40$.
The tension between the two objectives slows down learning (Figure~\ref{fig:svleffect}), but overfits less.
This gives better final test $S_n=S_p$ ($69.7\%$ and $69.6\%$ for depression and anxiety, respectively) and AUROC ($0.771$ and $0.768$).
\end{model}

Here paired inputs for the whisper backbone are constructed by concatenating all audio for a single voice, choosing a random $60$-second clip, and breaking this clip into two $30$-second inputs.
Where less than $60$ seconds of audio are available, minimally overlapping $30$-second clips are chosen instead.
In the case of Section~\ref{sec:or} without score variance loss, random $30$-second clips are chosen without any concatenation.

Since the SVL does not involve the label $y$, it also provides a method to incorporate unlabeled data.
Exploring SVL for semi-supervised learning in the context of this problem is left for future work.

\subsection{Knowledge distillation (KD)}
For patient privacy and consent reasons, we could only release models which take audio as input and do not perform transcription.
To maximize model accuracy within these constraints, we sought to transfer as much information as possible from text-input models into an audio-input model.
In this section we accomplish this with a knowledge distillation approach \citet{kd}. See Section~\ref{sec:llma} for another approach which combines synergistically with this.

The idea of knowledge distillation is to train a new network, called the \textbf{student}, to approximate the outputs of a fixed existing network, called the \textbf{teacher}.
Typically the student and teacher take the same input, with the teacher being a larger model with more resource-intensive inference and the student being a smaller model allowing cheaper inference.

The same framework applies when the student and teacher take related inputs with the same label, in this case audio and corresponding transcribed text data, with no restrictions on model size.
For a teacher of the form $\textrm{teacher}(x) = \textrm{teacher}_\textrm{text}(\textrm{transcription}(x))$, we train the student model parameters $\theta$ so $\textrm{student}_\theta(x) \approx \textrm{teacher}_\textrm{text}(\textrm{transcription}(x))$ for speech samples $x$.

The fine-grained details of the teacher output on the training set can provide a stronger training signal than the hard labels that were used for training the teacher.
A common intuition is that the teacher learned to make more nuanced distinctions than the labels provide directly, say by learning to assign a higher score $z_1 = \textrm{teacher}(x_1)$ to a particular labeled example $(x_1, y)$ than the score $z_2 = \textrm{teacher}(x_2)$ assigned to another example $(x_2, y)$ with the same label.
A student may learn better by mimicking these teacher scores than by trying to generate the labels directly.

This intuition is based on the common case that the label $y$ is a deterministic function of the input $x$.
In the present case this seems unlikely -- we do not expect $30$ seconds of speech on an arbitrary topic to definitively pin down a patient's PHQ-$9$ or GAD-$7$ scores.
In this case knowledge distillation can provide another advantage: while $y$ may not be a deterministic function of the input $x$, $\textrm{teacher}(x)$ is.
The teacher hides information about the labels which may not be predictable from the input, ensuring that the student's task would be solvable exactly with sufficient capacity.

\begin{model}[Teacher]
\label{model:kdteacher}
The best teacher model in the sense of test set performance was found to be a combined audio- and text-input model.
This is a parallel combination of Model~\ref{model:svl} on the audio and Model~\ref{model:text} on the transcription, adding the resulting scores, and subtracting the mean of the text scores over the training set.
The mean subtraction is to simplify the student task and does not affect the metrics.
These teacher scores yield a test $S_n=S_p$ of $71.0\%$ for depression and $70.8\%$ for anxiety, with AUCs of $0.789$ and $0.783$, respectively.
%
%
%
\end{model}

\begin{model}[Student]
\label{model:kdstudent}
The student was initialized to Model~\ref{model:or}, so at the start of training the teacher's scores on the training set were a zero-mean modification of the student's scores.
The student was then trained to minimize squared error between its scores and the teacher scores on the same training set.
SVL was applied with a relative weighting of $1:1$ between the student-teacher score squared error and the student score variance.
The student test $S_n=S_p$ was $70.3\%$ for depression and $70.2\%$ for anxiety, with AUCs $0.779$ and $0.773$, respectively.
See Figure~\ref{fig:svleffect} for the validation curve.
\end{model}

\subsection{Cross-modal distillation via LLM approximation}
\label{sec:llma}
Large Language Model (LLM) approximation can be viewed as a variant of knowledge distillation, with an important distinction:
rather than aligning the output logits of the teacher and student models, the objective is to align their intermediate representations.
Specifically, the student network is trained to match the embeddings produced by a teacher network (e.g., BERT).
Another key difference is that the student is implemented as a dedicated network, rather than reusing the primary audio backbone.
In this setting, an acoustic model is optimized to approximate the representational behavior of a language model as closely as possible.
In our case, this distillation is performed in a cross-modal setting, where the goal is to transfer depression- and anxiety-related biomarker
information from one modality (text) to another (audio).

\begin{figure}[tbp]
\begin{center}
\includegraphics[width=0.8\textwidth]{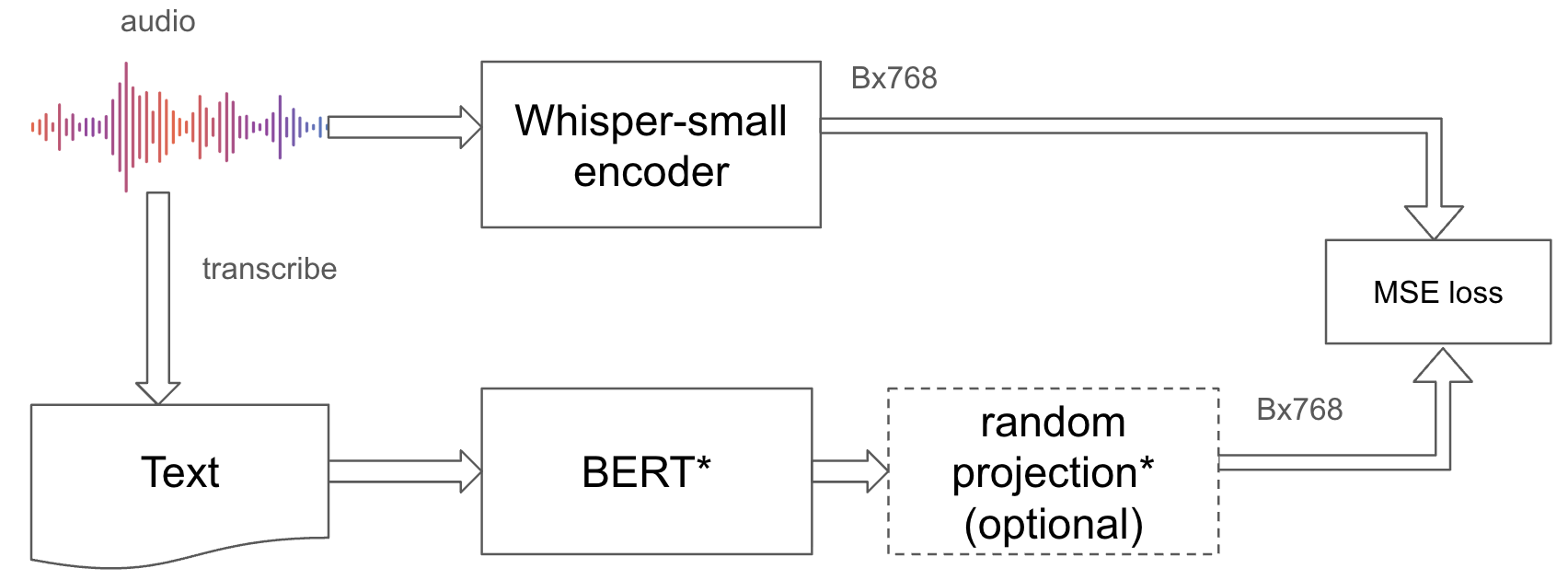}
\caption{Block diagram of LLM approxmimation process. Asterisk (*) means the neural net weights are frozen.}
\label{fig:llma_diagram}
\end{center}
\end{figure}

The block diagram of such a Large Language Model approximator (LLMA) is presented in Figure~\ref{fig:llma_diagram}.
It is worth noting that if the dimensionalities of the student and teacher embeddings do not match, the BERT embedding
can be passed through a random projection layer to ensure that both embeddings lie in the same space.
In our case, such a projection is not required, as \texttt{whisper-small.en} and \texttt{bert-base-uncased} produce embeddings of the same dimensionality (1 × 768).
This allows us to directly minimize the distance between the teacher and student representations.
During optimization, the teacher network and the projection layer (if present) remain frozen, ensuring that gradient updates affect only the LLMA model.
The loss functions used to train the LLMA include mean squared error (MSE) and cosine similarity between embedding vectors.
For unit-normalized vectors, minimizing MSE is equivalent to maximizing cosine similarity. However, in our case, the embeddings are not
normalized, and therefore these objectives lead to different optimization outcomes. Our experiments indicate that using MSE yields
better performance than cosine similarity.

\begin{figure}[tbp]
\begin{center}
\includegraphics[width=0.8\textwidth]{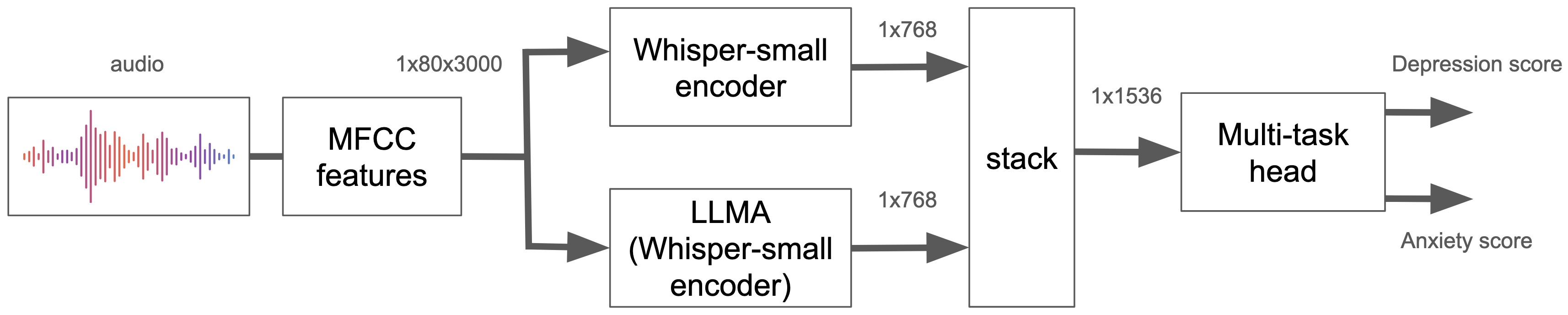}
\caption{DAM block diagram}
\label{fig:model_diagram}
\end{center}
\end{figure}

\begin{model}
\label{model:withllma}
\label{model:dam31}
The model is trained in the following multi-stage manner:
\begin{enumerate}
\item Fine-tuning the original biomarker encoder (Whisper Small) with score variance loss (please see Model~\ref{model:svl});
\item Fine-tuning the text-only Model~\ref{model:text} to produce ground-truth embedding vectors for LLM approximation;
\item LLM approximation stage;
\item Final head fine-tuning (see Figure~\ref{fig:model_diagram}).
\end{enumerate}
We use CORAL loss at all training stages, except LLM approximation stage where we leverage MSE loss. The recorded $S_n=S_p$ on the test was $71.1\%$ for depression and $70.7\%$ for anxiety, with AUCs $0.788$ and $0.781$, respectively.
\end{model}

During the final head fine-tuning stage, the objective is to train a multi-head classifier using embeddings from both backbones -- the
biomarker encoder and the LLMA. At this stage, the backbones are kept frozen. However, this is not a strict design choice, and in principle,
they may also be further optimized. In our experiments, keeping backbones frozen resulted in a significant training speedup
without causing measurable performance degradation.
We refer to this LLMA-powered system as \textbf{Model~\ref{model:withllma}} and its block diagram is presented in Figure~\ref{fig:model_diagram}.

\section{Discussion}
\label{sec:discussion}

In addition to our machine learning modeling efforts, we conducted a wide range of experiments,
including evaluating our models’ ability to extract biomarker-specific information from vocal characteristics rather than lexical content,
as well as their ability to distinguish between biomarkers associated with depression and anxiety.
In this section, we present a selection of these experiments.

\subsection{Separating effect of audio biomarkers and lingustic content}
\label{sec:audio_vs_lingustics}

In this experiment, we investigate how the performance of our classifier varies with changes in linguistic content.
This allows us to quantify the extent to which model performance is driven by depression- and anxiety-related acoustic biomarkers,
as opposed to linguistic cues present in the speech. To this end, we employ voice cloning as a mechanism for placing existing
speakers into diverse and controlled linguistic contexts.

To evaluate this approach, we use 590 utterances randomly sampled from our test.
Each utterance was voice-cloned with Sesame TTS, and the resulting synthetic voices were used to narrate a fixed passage from the novel \textit{Frankenstein},
ensuring identical linguistic content across both depressed and non-depressed speakers. This design isolates vocal biomarkers as the primary source of variation.
We then applied our Depression-Anxiety Model to the generated data, expecting sensitivity and specificity to exceed chance levels (\textasciitilde 50\%),
which would indicate successful transfer of depression-related characteristics.

The experiment was repeated using a second standardized passage from the science fiction novel \textit{1984} to assess consistency across different content.
Both passages are presented in the Appendix ~\ref{ref:app_books}.
\begin{table}[h]
\centering
\caption{Performance with voice cloning and altered linguistic content}
\begin{tabular}{lcc}
\hline
\textbf{Utterance contents} & $S_n=S_p$ & \textbf{AUC} \\
\hline
Original utterances (baseline) & 70.3 & 0.79 \\
Frankenstein & 61.9 & 0.69 \\
1984 & 61.8 & 0.67 \\
\hline
\end{tabular}
\label{tab:perf_voice_clonned}
\end{table}

The results of this experiment in the Table~\ref{tab:perf_voice_clonned} demonstrate that our classifier retains predictive capability even when linguistic content is controlled,
supporting the hypothesis that depression- and anxiety-related acoustic biomarkers play a meaningful role in model performance
While performance decreases relative to the baseline condition with original utterances (AUC 0.79), the model achieves consistent
above-chance performance on voice-cloned samples with standardized passages (AUC 0.69 for \textit{Frankenstein} and 0.67 for \textit{1984}).
This performance drop indicates that linguistic cues contribute substantially to classification accuracy;
however, the preservation of signal in the controlled setting confirms that non-linguistic vocal features carry discriminative information.
The similar results across two distinct passages further suggest that the observed effects are robust to variations in textual content.
It should be noted that our model has not been trained on synthetic data or on content with differing pronunciation styles and intonation (e.g., novels),
which may have partially contributed to the observed performance degradation.
Overall, these findings highlight that while linguistic information enhances model performance, acoustic biomarkers alone are sufficient
to support meaningful classification, reinforcing the importance of incorporating both modalities in depression and anxiety detection systems.

\subsection{Correlation between depression and anxiety scores}
\begin{figure}[tbp]
\begin{center}
\includegraphics[width=0.8\textwidth]{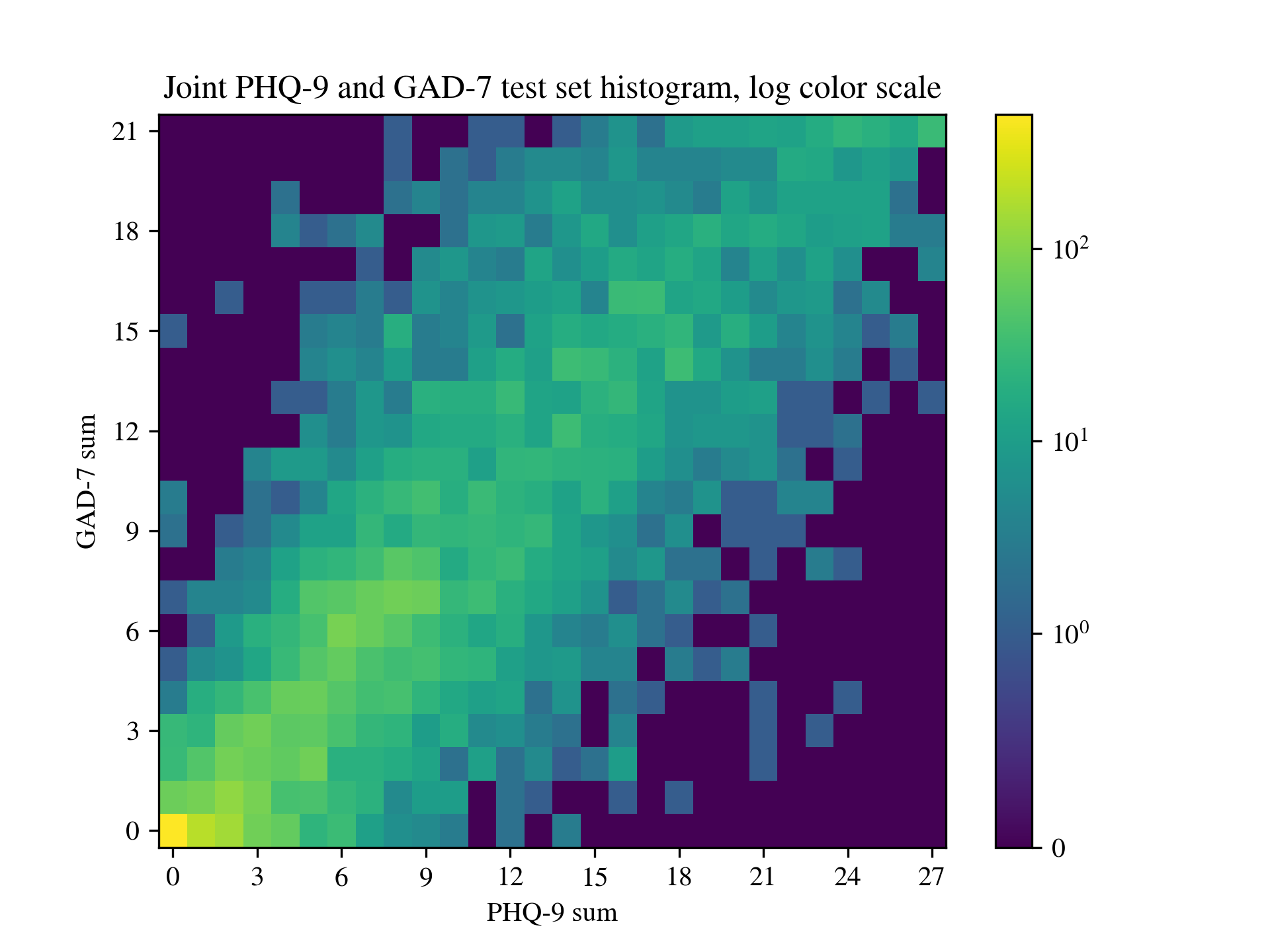}
\caption{Joint distribution of PHQ-$9$ and GAD-$7$ sums on the test set. Note that both are biased towards zero, so counts are represented on a log scale to show the correlation structure.}
\label{fig:phqgadcorr}
\end{center}
\end{figure}

\begin{figure}[tbp]
\begin{center}
\includegraphics[width=0.8\textwidth]{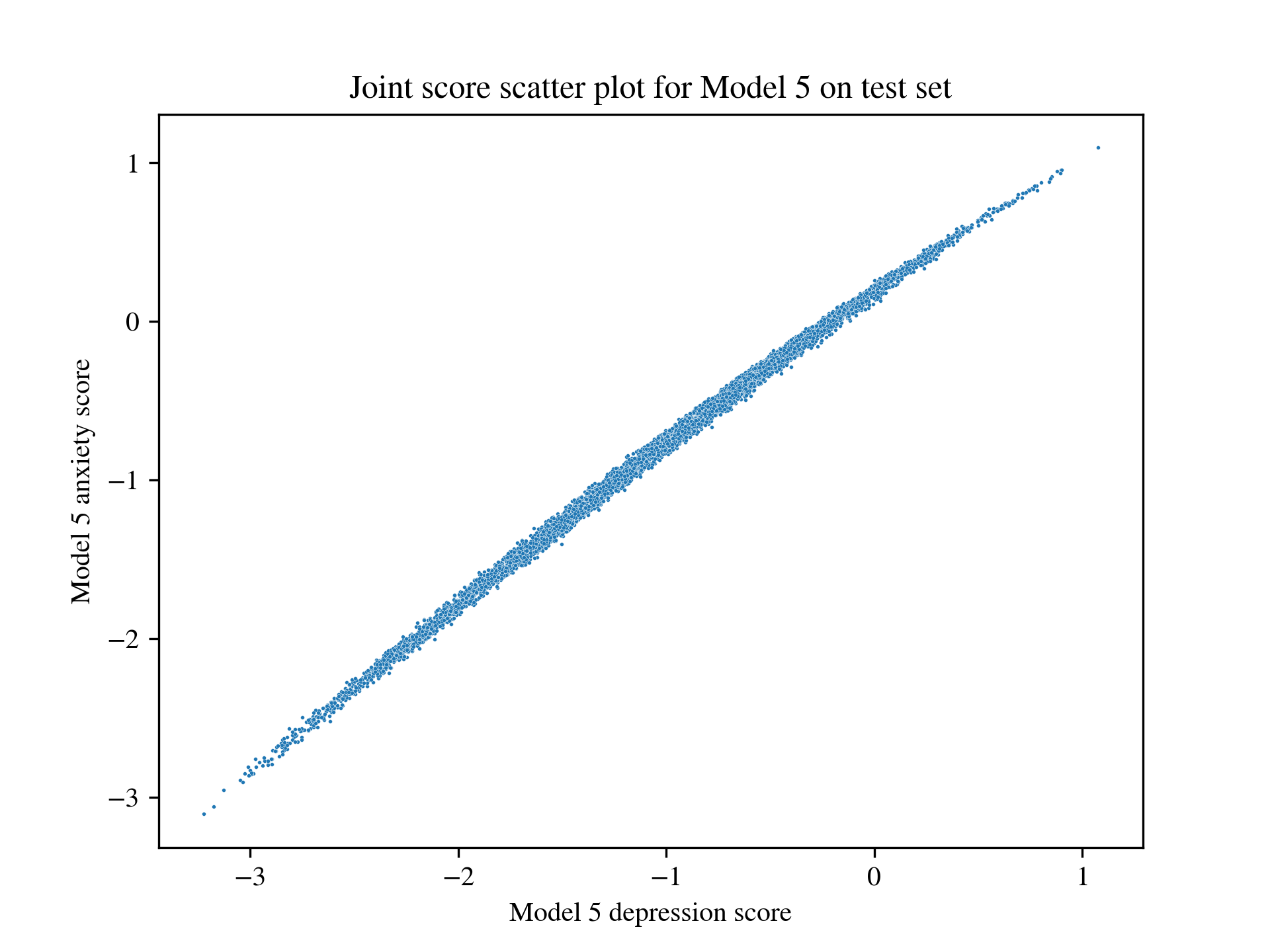}
\caption{Scatter plot of depression and anxiety scores from Model~\ref{model:dam31}. Note the almost perfect, almost linear correlation.}
\label{fig:scorecorr}
\end{center}
\end{figure}

Depression and anxiety are highly comorbid \citet{hirschfeld:comorbid, kessler:comorbid, kessler:who}, which here manifests as the PHQ-$9$ and GAD-$7$ sums being correlated (Figure~\ref{fig:phqgadcorr}).
One might expect the depression and anxiety model scores to be similarly correlated, but in fact they are much more highly correlated (Figure~\ref{fig:scorecorr}). Linear and ordinal correlation measures are shown in Table~\ref{tab:corr}.

It remains an open question why the scores are so much more highly correlated than the corresponding labels.

\begin{table}[!ht]
\caption{Correlation measures for depression and anxiety labels compared to scores.}
\label{tab:corr}
\small
    \centering
    
\begin{tabular}{lrrr}
\toprule
& Pearson $r$ & Spearman $\rho$ & Kendall $\tau$ \\
\midrule
PHQ-$9$ sum vs.\ GAD-$7$ sum &
0.846 &
0.852 &
0.698 \\
Model~\ref{model:dam31} depression vs.\ anxiety scores &
0.999 &
0.999 &
0.970 \\            
\bottomrule
\end{tabular}

\end{table}

\section{Conclusions}
\label{sec:conclusions}

We have presented a model to predict the severity of depression and anxiety from 30 seconds of speech. We experimented with multiple different modeling techniques, including studying various pre-trained backbones, loss functions, and knowledge distillation, to consistently improve the model's performance over the iterations. The culmination of these techniques resulted in the model achieving $71.1$\% depression $S_n=S_p$ and $70.7$\% anxiety $S_n=S_p$. We believe the model can serve as a foundation model for detecting depression and anxiety in speech, and the descriptions provided in this paper can help guide future research direction. We have open-sourced the model at \url{https://huggingface.co/KintsugiHealth/dam}].

\section*{Acknowledgements}
\label{sec:acknowledgements}
The research presented in this paper was conducted by Oleksii Abramenko, Noah Stein, and Colin Vaz during their work at Kintsugi Mindful Wellness Inc.
It builds upon years of prior modeling, data collection, clinical research, and operational efforts contributed by a broader team.
The open-source Depression-Anxiety Model, along with a full list of contributors, is available at \href{https://huggingface.co/KintsugiHealth}{https://huggingface.co/KintsugiHealth}.
This paper is based solely on publicly available materials released by Kintsugi and does not reference any confidential information.

{
\small

\bibliographystyle{plainnat}
\bibliography{references}
}

\clearpage
\appendix
\refstepcounter{section}
\section*{Appendix \thesection\ -- Passages used for synthetic data generation (see Section~\ref{sec:audio_vs_lingustics})}
\label{ref:app_books}

\textbf{"Frankenstein" passage [Mary Shelley 1818 -- Vol.\ I, Letter I]}

\begin{quote}
Six years have passed since I resolved on my present undertaking. I can, even now, remember the hour from which I dedicated
myself to this great enterprise. I commenced by inuring my body to hardship. I accompanied the whale-fishers on several
expeditions to the North Sea; I voluntarily endured cold, famine, thirst, and want of sleep; I often worked harder than
the common sailors during the day and devoted my nights to the study of mathematics, the theory of medicine, and those branches of physical
science from which a naval adventurer might derive the greatest practical advantage.
\end{quote}

\textbf{"1984" passage [George Orwell 1949 -- Part I, Section I]}

\begin{quote}
The Ministry of Truth contained, it was said, three thousand rooms above ground level, and corresponding ramifications below.
Scattered about London there were just three other buildings of similar appearance and size. So completely did they dwarf
the surrounding architecture that from the roof of Victory Mansions you could see all four of them simultaneously.
They were the homes of the four Ministries between which the entire apparatus of government was divided.
The Ministry of Truth, which concerned itself with news, entertainment, education, and the fine arts. The Ministry of
Peace, which concerned itself with war.
\end{quote}

\end{document}